\documentclass[runningheads]{llncs}
\usepackage[T1]{fontenc}
\usepackage{cite}
\usepackage{amsmath,amssymb,amsfonts}
\usepackage{algorithmic}
\usepackage{graphicx}
\usepackage{textcomp}
\usepackage{multirow}
\usepackage{comment}
\usepackage{amsmath}
\usepackage{algorithm}
\usepackage{xcolor}

\begin{document}
\title{Benchmarking histopathology foundation models in a multi-center dataset for skin cancer subtyping}
\titlerunning{Benchmarking histopathology FM}
%
\author{Pablo Meseguer\inst{1}$^*$\orcidID{0000-0001-7821-6168} \and
Rocío del Amorr\inst{1,2}\orcidID{0000-0002-5342-2093} \and
Valery Naranjo\inst{1,2}\orcidID{0000-0002-0181-3412}}
\authorrunning{P. Meseguer et al.}
%


\institute{Universitat Politècnica de València (UPV) \\ Valencia, Spain  \and
Artikode Intelligence S.L. \\ $^*$ Corresponding author: Pablo Meseguer (\email{pabmees@upv.es})}

\maketitle              
\begin{abstract}
\sloppy
Pretraining on large-scale, in-domain datasets grants histopathology foundation models (FM) the ability to learn task-agnostic data representations, enhancing transfer learning on downstream tasks. In computational pathology, automated whole slide image analysis requires multiple instance learning (MIL) frameworks due to the gigapixel scale of the slides. The diversity among histopathology FMs has highlighted the need to design real-world challenges for evaluating their effectiveness. To bridge this gap, our work presents a novel benchmark for evaluating histopathology FMs as patch-level feature extractors within a MIL classification framework. For that purpose, we leverage the AI4SkIN dataset, a multi-center cohort encompassing slides with challenging cutaneous spindle cell neoplasm subtypes. We also define the Foundation Model - Silhouette Index (FM-SI), a novel metric to measure model consistency against distribution shifts. Our experimentation shows that extracting less biased features enhances classification performance, especially in similarity-based MIL classifiers. 

\end{abstract}
\section{Introduction}
Computational pathology (CPath) has recently experienced an overwhelming transformation due to foundation model (FM) development. The emergence of self-supervised pretraining strategies and high-capacity neural networks has permitted training general-purpose models with enhanced representation learning \cite{dosovitskiy2020image,oquab2024dinov2}. These breakthroughs in artificial intelligence, coupled with the availability of multiple large-scale datasets of histopathological imaging, have driven the development of a remarkable collection of state-of-the-art histopathology FMs. These models promise to effectively tackle diverse downstream CPath tasks ranging from visual applications such as patch-level classification to multi-modal challenges like tissue captioning. 

Histopathology FMs are characterized through a combination of the pretraining paradigm, the assembled dataset, and the model architecture. The need for massive datasets to train large models has turned the focus to less restrictive forms of supervision. In particular, we differentiate between vision-only self-supervised strategies based on masked image modeling \cite{oquab2024dinov2} and vision-language supervision \cite{radford2021learning} to learn visual features from text supervision. Although each model uses slightly different configurations for model scale, vision transformers (ViT) \cite{dosovitskiy2020image} have emerged as the most popular choice for model architecture. Histopathology FMs primarily differ in the size, diversity, and quality of their pretraining corpora, which collection is conditioned by the pretraining strategy. While self-supervised pretraining relies on millions of patches extracted from massive collections of slides, vision-language supervision requires histopathology images paired with their textual description.

Automated whole slide imaging (WSI) analysis is the most promising challenge in CPath, as it mimics the diagnostic process of pathologists. Specialized scanners digitize tissue samples into WSI, enabling their incorporation into computer vision applications. However, the gigapixel scale of the slides makes them unmanageable by current hardware and leads to the incorporation of multiple instance learning (MIL) paradigms, a particular form of weakly-supervised learning. In MIL paradigms for CPath, a bag (slide) comprises multiple instances (patches), and the model aims to predict the bag-level label.  

The diversity among pretraining strategies and evaluated downstream tasks has resulted in a heterogeneous comparison of histopathology FMs. Despite initial efforts on benchmarking, authors in \cite{campanella2024clinical} solely focused on self-supervised models and relied on single-center cohorts, which hinders insights about model consistency when facing distribution shifts. To solve this gap, this work comprehensively evaluates histopathology FMs pretrained under different forms of supervision in a skin cancer subtyping task. In particular, we refer to the AI4SkIN dataset \cite{del2025fusocelular}, which encompasses WSIs of six cutaneous spindle cell (CSC) neoplasms, to evaluate the downstream performance in a MIL-based slide-level classification task. Additionally, we present the Foundation Model - Silhouette Index (FM-SI) to measure model robustness against scanner shifts and how it can limit the proficiency of the models in real-world scenarios. Overall, the proposed benchmark provides a robust assessment of model generalization and reliability in a challenging clinical task. The main contributions of our work can be summarized as follows:

\begin{itemize}
    \item We conduct a comprehensive evaluation of histopathology foundation models, benchmarking their performance on a challenging, multi-center whole slide image dataset.
    \item We assess the effectiveness of each foundation model using two distinct MIL strategies, offering insights into their adaptability performance.
    \item We introduce the FM-SI metric to quantify the center-related information leakage in the extracted feature representations extracted, enabling the assessment of model consistency against distribution shifts.
\end{itemize}

\section{Methodology}
\begin{figure*}[h!]
\centerline{\includegraphics[width=\textwidth]{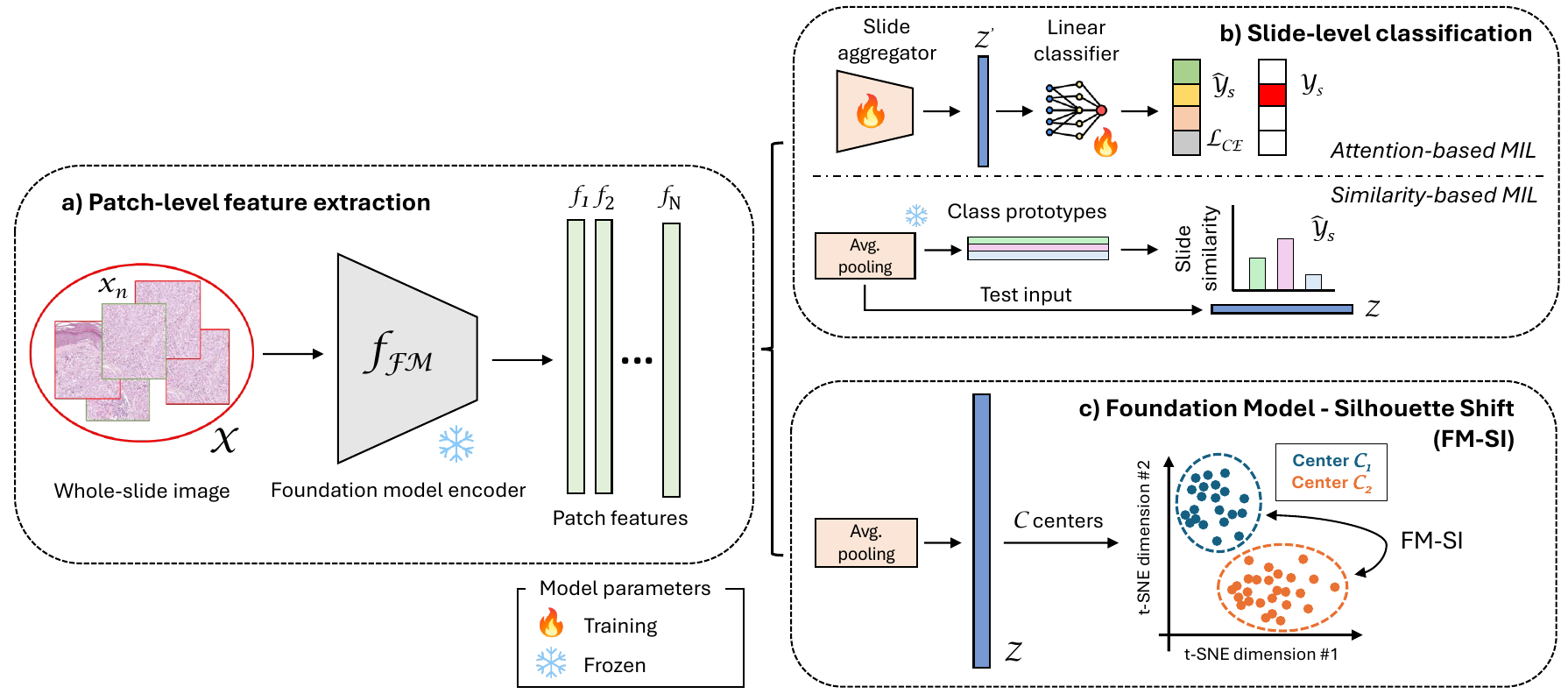}}
\caption{
\textit{Histopathology foundation model evaluation framework}: Given a whole slide image ($X$), the feature extraction stage uses a foundation model encoder ($f_{FM}$) to extract patch representations ($f_n$). To address the slide-level downstream classification task, we rely on attention-based MIL (ABMIL) \cite{ilse2018attention} including a slide aggregator and linear classifier optimized through the categorical cross-entropy loss ($\mathcal{L_{CE}}$) and a similarity-based MIL classifier (MI-SimpleShot) \cite{chen2024towards} which relies on class prototypes to compute the slide-level prediction ($\hat{Y}_s$). Finally, the Foundation Model - Silhouette Shift (FM-SI) for the $C$ available centers in the dataset is measured on top of the slide embeddings ($Z$) to evaluate the presence of scanner-related information in the extracted features.
}
\label{fig_framework}
\end{figure*}

This section includes a description of the methodology used for the evaluation of histopathology FMs including the problem formulation, a description of the models and details on the MIL settings for slide classification. A graphical framework is provided in Figure \ref{fig_framework}.

\subsection{Foundation model evaluation framework}

\sloppy
\subsubsection*{Problem formulation:} Computational pathology challenges manage the gigapixel scale of WSI by dividing whole slides into patches such as $X = \{x_n\}_{n=1}^{N}$, where $N$ denotes the number of instances in a bag. In the multi-class scenario, each WSI is a member of one of $S$ mutually exclusive classes, being $Y_{s} \in \{0, 1\}$ the slide-level ground truth. In a multi-center dataset, each WSI is associated with the pathology department of a particular hospital ($C$) from which it was acquired and digitized. In a slide classification task under the MIL paradigm, we aim to train a model capable of predicting bag-level labels using a combination of features extracted at the instance level by a histopathology FM. Let us denote a histopathology FM encoder ($f_{FM}$) which projects patches to a lower dimensional manifold $f \in \mathcal{F} \subset \mathbb R^{d}$, where $d$ refers to the embedding dimension.

\subsubsection{Histopathology foundation models:}
Large-scale screening programs and the expansion of digital pathology have permitted the collection of massive datasets of histopathological imaging, thus accelerating the development of foundation models for this particular medical imaging modality. This work considers up to 8 histopathology FM trained under different forms of supervision to operate as patch-level feature extractors in a MIL-based classification framework, as shown in Figure \ref{fig_framework} a). Table \ref{tbl_models} summarizes each histopathology FM regarding its pretraining strategy, the embedding dimension, the number of model parameters, and the size of the pretraining dataset. Our comparison covers from vision-only models pre-trained with self-supervised strategies like UNI \cite{chen2024towards}, VIRCHOW-2 \cite{zimmermann2024virchow2}, GPFM \cite{ma2024towards}, and CHIEF \cite{wang2024pathology} to multimodal vision-language models such as CONCH \cite{lu2024visual}, MUSK \cite{xiang2025vision}, KEEP \cite{zhou2024knowledge}, and PLIP \cite{huang2023visual}. 

It is important to highlight the unique configuration of each model. Notably, the length of the latent space ($d$) varies between more condensed vectors for CONCH and PLIP ($d=512$) and larger feature representations like MUSK ($d=2048$). This configuration is critical as it will condition the models' size for a particular downstream task. Although ViTs are transversal to all FMs, we also note that ViT-Large with a patch size equal to 16 (ViT-L/16, 303 million parameters) is the most popular encoder as selected by UNI, MUSK and GPFM models.

Regarding dataset size, vision-language models are generally trained with fewer histopathology images, as language-image pretraining supposes a more restrictive form of supervision. As an exception, MUSK \cite{zhou2024knowledge} was trained with 50 million (M) images with masked image modeling and later used one million text reports for multimodal alignment. VIRCHOW-2 was pre-trained on tiles extracted from 3.1 million slides, making it the largest training corpus among all models, marking a notable leap to the next largest dataset.   

\begin{table}[htpb]
    \caption{A summary of the selected histopathology foundation models}
    \centering
    \setlength{\tabcolsep}{7pt}
    \renewcommand{\arraystretch}{1.05}
    \begin{tabular}{lrrrr}

        \hline
        \textbf{Model} & {\textbf{\begin{tabular}[l]{@{}r@{}}Pretraining \\ strategy\end{tabular}}} & 
        \textbf{\begin{tabular}[l]{@{}r@{}}Feature \\ length (d)\end{tabular}} & 
        \textbf{\begin{tabular}[l]{@{}r@{}}Params. \\ (M)\end{tabular}} & 
        \textbf{\begin{tabular}[l]{@{}r@{}}Tiles \\ (M)\end{tabular}} \\
        \hline
        UNI \cite{chen2024towards}        & \multirow{4}{*}{\begin{tabular}[r]{@{}r@{}}Self supervision\end{tabular}}  & 1024  & 303 & 100   \\
        VIRCHOW-2 \cite{zimmermann2024virchow2} &  & 1280  & 632  & 3.1*  \\
        GPFM \cite{ma2024towards}         &  & 1024  & 303  & 190   \\
        CHIEF \cite{wang2024pathology}    &  & 768   & 28   & 15    \\\hline
        CONCH \cite{lu2024visual}         & \multirow{4}{*}{\begin{tabular}[r]{@{}r@{}}Vision-language\\ supervision\end{tabular}} & 512   & 90   & 1.7   \\
        MUSK \cite{xiang2025vision}       &  & 2048  & 303  & 50    \\
        KEEP \cite{zhou2024knowledge}     &  & 768   & 414  & 1     \\
        PLIP \cite{huang2023visual}       &  & 512   & 87   & 0.21  \\\hline
    \end{tabular}
    \label{tbl_models}
    \parbox{\linewidth}{\vspace{0.5mm} \raggedleft \textit{Note}: * denotes number of slides}
\end{table}

\subsubsection{Multiple instance learning (MIL) classifiers:}
As presented in Figure \ref{fig_framework} b), we used two MIL approaches to assess the performance of each foundation model in a slide classification task. 

\paragraph{Attention-based multiple instance learning (ABMIL) \cite{ilse2018attention}:}
We utilized the attention-based MIL to train a slide aggregator on top of the features extracted by the foundation model. ABMIL processes patch features through a learnable attention layer, which assigns attention weights to each patch reflecting its significance in the classification. Finally, a weighted average of instance features based on the learned scores outputs a slide-level representation ($Z'$) of the slide, which is forwarded to a linear classifier to compute the bag-level predicted probabilities ($\hat{Y}_s$). Model parameters of the slide aggregator and the classification layer are updated by gradient descent using the categorical cross-entropy ($\mathcal{L_{CE}}$) between the slide ground truths ($Y_s$) and predictions ($\hat{Y}_s$) as the cost function. 

\sloppy
\paragraph{Similarity-based classifier (MI-SimpleShot) \cite{chen2024towards}:}
Multiple Instance (MI) - SimpleShot is a non-parametric approach for slide-level classification in CPath. Given a set of training data in the form of mean pooled feature embeddings ($Z$) representing the $S$ classes in the task at hand, MI-SimpleShot constructs the s-th class prototype as the centroid of the feature representations for the data points of the s-th class within the training subset. For classification, it computes the slide-level cosine similarity between the constructed prototypes and the test input query to determine the slide's prediction ($\hat{Y}_s$). This approach is advantageous for evaluating the ability of pretrained feature extractors to generate robust feature representation without requiring additional training.
\par

\subsection{Center shift measurement}
Various sources of distribution shift, such as differences in demographics, acquisition centers, staining protocols, and digital scanners, can significantly impact CPath tasks, potentially limiting model performance in real-world applications. Despite the large-scale pretraining stage of histopathology foundation models promising to obtain robust representation independent of these noise sources, careful evaluation of model consistency against domain shifts is required. This work focuses on measuring center-related scanner shifts as two scanners were used to digitize the tissue samples on the benchmarking dataset. 

Inspired by intuitive data visualizations of high-dimensional feature and clustering analysis, we propose \textbf{F}oundation \textbf{M}odel \textbf{S}ilhouette \textbf{I}ndex (FM-SI) (see Algorithm \ref{algorithm_metric}), a novel metric to measure the center-related shift of histopathology FM. Given the dataset ($D$) of a downstream task composed of a set of patches for each slide ($X$), we initially extract the patch-level features using a particular foundation model ($f_{FM}$) and compute the slide embedding ($Z$) by mean pooling all instance features. In the following, we reduce the feature dimensionality of $Z$ to two components using t-Distributed Stochastic Neighbor Embedding (t-SNE) \cite{van2008visualizing} by minimizing the Kullback-Leibler divergence between the similarities of the high- and low-dimensional data. Finally, the silhouette coefficient \cite{rousseeuw1987silhouettes} is calculated on top of the 2D slide representation by measuring the mean intra-cluster distance and the mean nearest-cluster distance, considering as clusters the different centers ($C$) in the dataset. An FM-SI score closer to 0 means overlapping clusters, while a higher values closer to 1 translates in better clustered data points according to the center label, thus denoting more center-biased features. 

\begin{algorithm}[h]
\caption{Foundation Model - Silhouette Index (FM-SI)}
\label{algorithm_metric}
\begin{algorithmic}[1]
\STATE \textbf{Inputs:} Dataset \( D = \{X_i\} \), center labels ($C_i$) and a foundation model ($f_{FM}$)
\STATE \textbf{Output:} FM-SI score ($\sigma$)
\STATE \textbf{Step 1}: Obtain the slide representation (Z)
\FOR{each slide \( i \in D \)}
    \STATE $f_n = f_{FM}(x_n)$ \COMMENT{Extract patch features}
    \STATE \( Z = \frac{1}{N} \sum_{n=1}^{N} f_{_n} \) \COMMENT{Compute slide embedding}
\ENDFOR
\STATE \textbf{Step 2}: 2D t-SNE dimensionality reduction on \( Z \)
\STATE \hspace{.25cm} $Z_{2D} = \text{t-SNE}(Z)$ \COMMENT{following \cite{van2008visualizing}}
\STATE \textbf{Step 3}: Compute silhouette score $\sigma$
\STATE \hspace{.25cm} $\sigma = \text{SilhouetteScore}(Z_{2D}, C)$ \COMMENT{following \cite{rousseeuw1987silhouettes}}
\STATE \textbf{Return} \( \sigma \) as the FM-SI score
\end{algorithmic}
\end{algorithm}

\section{Experimental configuration}
In this section, we provide a summary of the experimental settings and present the multi-center dataset to define the benchmark of the downstream task. 

\subsection{Experimental settings}
We follow a 5-fold stratified cross-validation partition to validate both MIL approaches. We measure the balanced accuracy (BACC) score as the average of recall scores (sensitivity) across all classes, which is particularly useful for class-imbalanced datasets. Regarding the MIL approaches, attention MIL models were trained for 20 epochs with an AdamW, a cosine learning rate scheduler, and a peak learning rate of $10^{-4}$. Weighted cross-entropy loss was used to consider less-represented classes fairly. We set the size of the intermediate layer in the attention network in ABMIL to a quarter of the input size of the feature to handle the different sizes of the latent space. Note that the similarity-based classifier does not require parameter updates. MI-SimpleShot constructs the prototypes directly from the training samples and computes the slide-level cosine similarity for a given test query without requiring parameter update.

\subsection{AI4SkIN dataset}
The AI4SkIN dataset \cite{del2025fusocelular} encompasses WSIs from six subtypes of CSC neoplasms: leiomyomas (lm), leiomyosarcomas (lms), dermatofibromas (df), dermatofibrosarcomas (dfs), spindle cell melanomas (scm), atypical fibroxanthomas (afx). CSC neoplasms are skin lesions ranging from benign to malign tumors with notable morphological overlaps that are supposed to be a diagnostic challenge even for experienced pathologists. The subtle differences make it challenging for computer vision systems to discover hidden patterns for subtype identification. The emergence of histopathology FMs promises to extract discriminative features that may enhance the performance of computer-aided diagnosis systems for skin cancer subtyping.

The dataset is constituted of whole-slide images obtained by digitizing tissue sections of biopsies resected from patients with a skin cancer diagnosis. In particular, WSIs were digitized at two different centers in Spain: Hospital Clínico Universitario de Valencia (HCUV) and Hospital Universitario San Secilio (HUSC) in Granada. The digitization process was done at 40x equivalent magnification with Roche’s scanner (Ventana iScan HT) at HCUV and Philips Ultra Fast Scanner at HUSC. The different scanners used for data acquisition generate distribution shifts that require careful consideration given its potential impact in real-world clinical applications, potentially leading to a decline in model performance. For that purpose, we propose a novel metric for distribution shift measurement and analyze its effect in the classification task.

We present an overview of AI4SkIN dataset in Table \ref{tbl_dataset}, including the counts of WSI for each center and skin cancer subtype. The proposed benchmark dataset consists of 621 slides, with 42.8\% coming from the hospital in Valencia (HCUV) and 58.2\% from the hospital in Granada (HUSC). It is worth mentioning that malignant subtypes (leiomyosarcoma and dermatofibrosarcoma) are underrepresented compared to its benign forms (leiomyoma and dermatofibroma) containing at least 2x more data samples. 

\begin{table}[h]
\caption{Counts of slides the AI4SkIN dataset per subtype and center}
\setlength{\tabcolsep}{7pt}
\renewcommand{\arraystretch}{1.05}
\begin{center}
    \begin{tabular}{lccc}
    \hline
    \textbf{Skin cancer subtype}    & \textbf{HCUV} & \textbf{HUSC} & \textbf{Overall} \\ \hline
    Leiomyoma              & 31   & 73   & 104     \\
    Leiomyosarcoma        & 21   & 23   & 44      \\ 
    Dermatofibroma        & 101  & 93   & 194     \\ 
    Dermatofibrosarcoma   & 21   & 34   & 55      \\ 
    Spindle cell melanoma & 48   & 74   & 122     \\ 
    Atypical fibroxanthoma & 44   & 58   & 102     \\ \hline
    \textbf{Overall}                & \textbf{266}  & \textbf{355}  & \textbf{621}     \\ \hline
    \end{tabular}
\label{tbl_dataset}
\end{center}
\end{table}

\section{Results}
This section evaluates the presence of center-related information in the representations extracted by the foundation models and examines its impact on the performance of downstream MIL models.

\subsection{Center shift measurement}
Careful evaluation of model consistency against distribution shifts is fundamental for the integration of CPath applications in clinical practice. Recently, \cite{de2025current} proposed the Robustness Index (RI), a new metric to measure how much biological features dominate over confounding factors in pathology FM. The RI computes a ratio based on the proportion that belongs to the same biological class versus those that belong to the same medical center. In contrast, FM-SI provides a measurement of center shift without requiring class labels that intuitively correlates with the representation of the slide-level latent space. 

\begin{figure}[h]
\centerline{\includegraphics[width=0.85\textwidth]{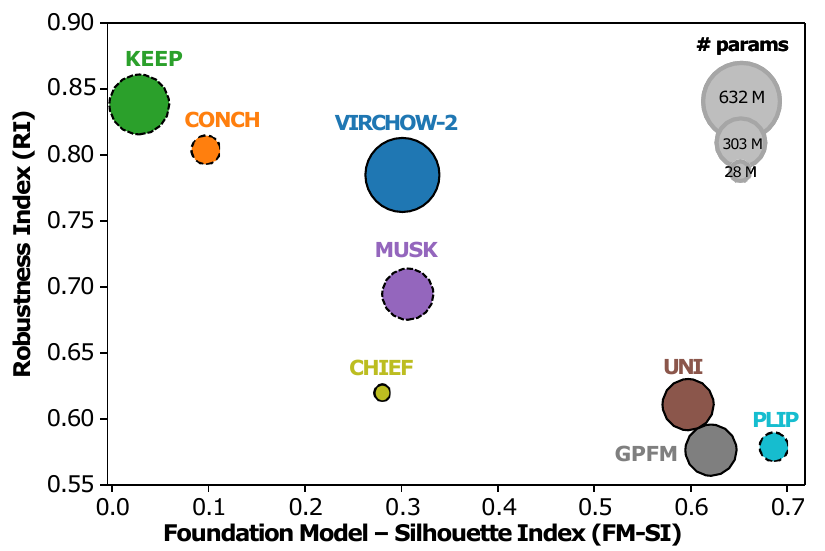}}
\caption{Measurement of center shift consistency of histopathology foundation models in terms of Robustness Index (RI) \cite{de2025current} and Foundation Model - Silhouette Index (FM-SI).}
\label{fig_metrics}
\end{figure}

Initially, we show in Figure \ref{fig_metrics} how the Foundation Model - Silhouette Index correlates with the Robustness Index \cite{de2025current}. Our metric assesses how closely data points from each center cluster together, with higher values suggesting that the data representations extracted by a histopathological FM capture significant center-related information. Therefore, lower FM-SI values mean higher preponderance of pathological features and correspond with higher values of the robustness score for a particular model. Overall, FM-SI shows a strong correlation with the RI with a correlation coefficient of $|\rho| = 0.890$. 

\begin{figure}[h!]
\centerline{\includegraphics[width=0.685\textwidth]{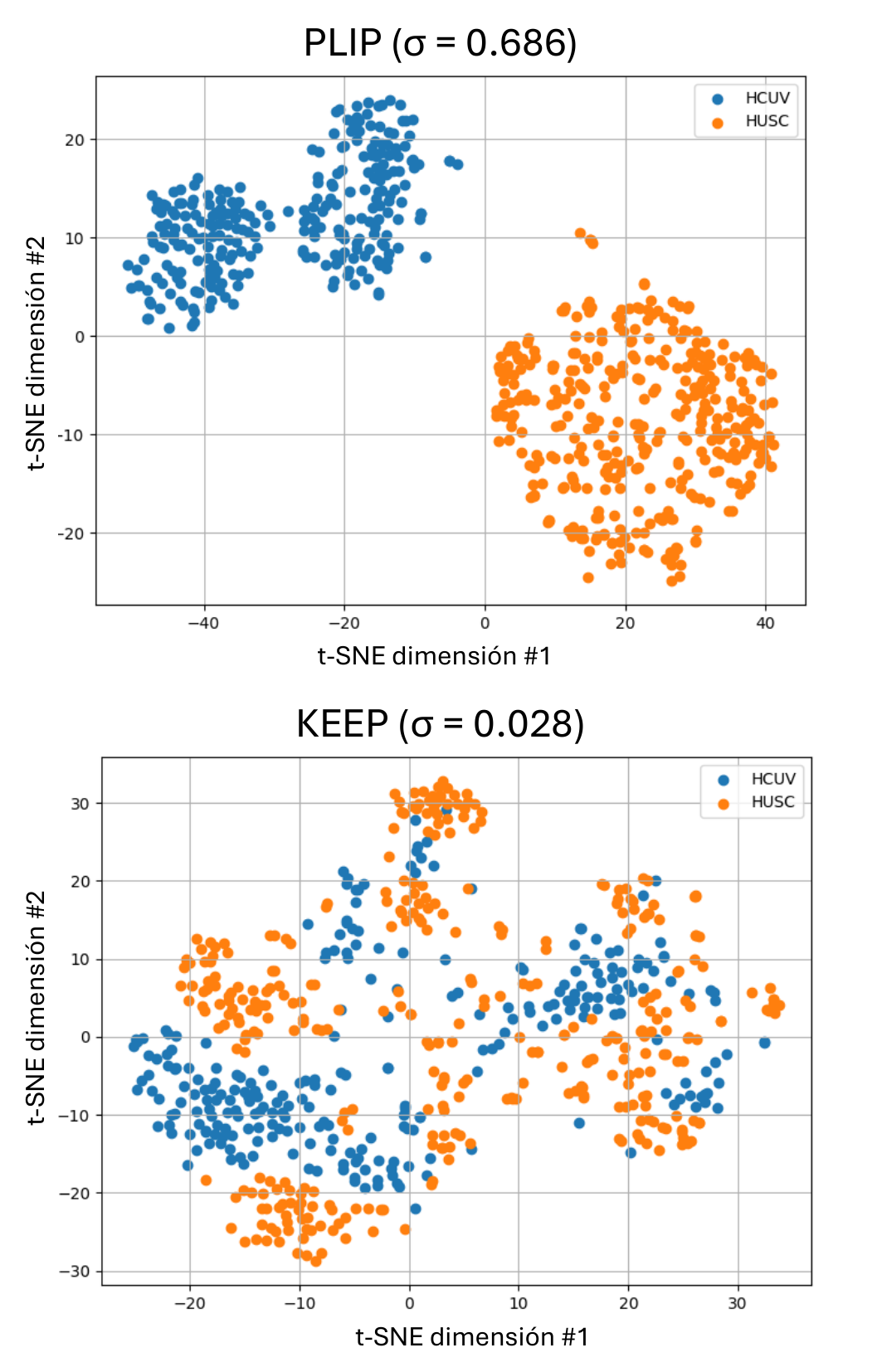}}
\caption{Latent space visualization at the slide-level of two histopathology foundation models: PLIP \cite{huang2023visual} and KEEP \cite{zhou2024knowledge}.}
\label{fig_tsne}
\end{figure}

Additionally, Figure \ref{fig_tsne} includes the 2D t-SNE \cite{van2008visualizing} scatter plots where each point corresponds to the slide-level features ($Z$) extracted either by the PLIP or KEEP models and the color indicates the center where the sample was acquired. These models were selected as they return the highest (PLIP with $\sigma\ = 0.686$) and the lowest (KEEP with $\sigma = 0.028$) values for the FM-SI metric. The analysis of the latent space shows that the PLIP encoder strongly clusters the samples according to each center, while data points are more sparse in terms of acquisition center for the KEEP model. This illustration is a clear example of how the proposed metric intuitively correlates with dimensionality reduction techniques commonly used for exploratory data analysis of deep features. 

\subsection{Classification performance}
\begin{figure}[h!]
\centerline{\includegraphics[width=0.915\textwidth]{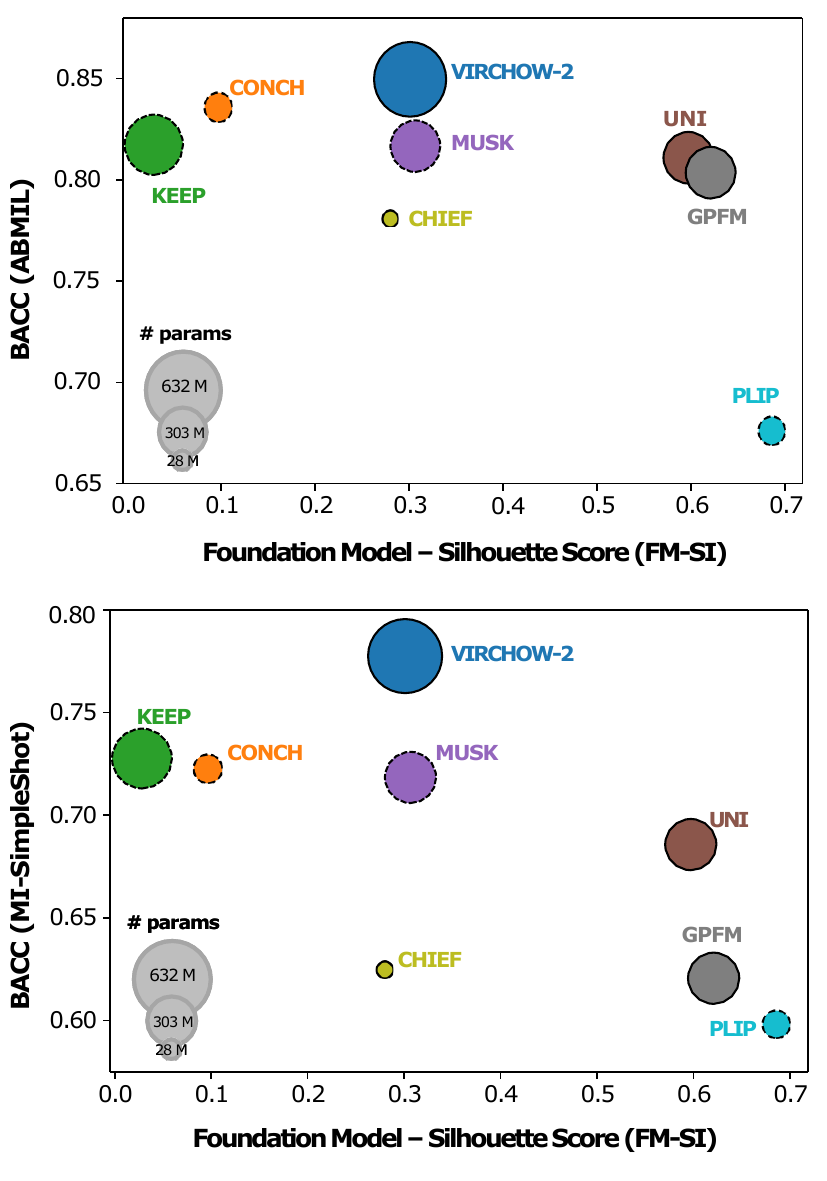}}
\caption{Downstream performance in terms of balanced accuracy (BACC) for skin cancer subtyping on the AI4SkIN dataset with ABMIL (top) and MI-SimpleShot (bottom) correlated with the Foundation Model - Silhouette Index (FM-SI) metric. Solid and dotted lines represent self-supervised and vision-language models, respectively.}
\label{fig_MIL}
\end{figure}

As previously stated, we investigate two different MIL approaches to address the slide-level classification task. We compare a MIL model relying on attention weights to obtain the slide representation embedding (ABMIL) and a non-parametric method (MI-SimpleShot) that computes similarities between constructed class prototypes and test input queries. To assess the potential impact of inter-center variability in the proposed benchmark, we analyze how the Foundation Model - Silhouette Index affects the downstream proficiency of FM for each MIL approach. For that purpose, we show in Figure \ref{fig_MIL} a scatter chart where we plot the performance of each MIL approach and the FM-SI of the model. 

It is worth noting than MI-SimpleShot performance presents a higher dependence on the center-biased representations as the trend shows a correlation between lower performance and higher values of the FM-SI metric ($R^2 = 0.428$). ABMIL improved and less-dependent effectiveness ($R^2 = 0.346$) indicates that learning phase enables to discover the relevant features and instances to enhance the subtyping classification compared to the non-parametric approach of MI-SimpleShot. Overall, attention-based MIL outperforms the MI-SimpleShot classifier by an average of 11.88\% across all evaluated encoders. This decrease is particularly critical in GPFM, which outputs 80.88\% balanced accuracy with ABMIL, but just 62.13\% with MI-SimpleShot.

We find that VIRCHOW-2 excels in both similarity- and attention-based MIL classifiers reaching 77.75\% and 86.81\% balanced accuracy, respectively. Despite VIRCHOW-2 outputs the fourth lowest FM-SI value, it surpasses the next performing model by 4.92\% on MI-SimpleShot and 3.81\% in ABMIL. Since VIRCHOW-2 is pre-trained on the largest dataset and uses the most complex model, these findings emphasize the impact of scaling laws on corpus size and model weights in histopathology FM. However, the VIRCHOW-2 pretraining dataset was assembled with 85\% of the slides from the same medical institution \cite{zimmermann2024virchow2} highlighting that dataset diversity is crucial to obtain robust center-independent data representations. 

Furthermore, PLIP gets the lower proficiency for both MIL methods with a notable reduction in attention-based classification (67.53\%), more than ten points lower than the second-bottom performer (CHIEF). The fact that PLIP collected pathology image-text pairs from Twitter raises critical questions about the quality and source knowledge of the data for pretraining FM. It is also worth highlighting that CONCH (90M params.) promotes models efficiency and reduces computational costs as it obtains comparable performance to MUSK and KEEP (303M params) which require model architectures with a notably higher number of parameters. Moreover, CONCH and KEEP encoders extract the few center-biased features suggesting that the incorporation of vision-language supervision to the pretraining stage provides them a higher focus on biological features self-supervised strategies relying solely on visual information. 

\section{Conclusions}
The continuous development of histopathology foundation models remarks the need for defining benchmarks to validate the performance of these models. This work presents a novel benchmark through the AI4SkIN dataset to compare self-supervised and vision-language histopathology foundation models in a challenging skin cancer subtyping task. Through two different multiple instance learning approaches, we evaluated the ability of FMs to extract discriminative features for addressing the downstream task while measuring the impact of center bias in the extracted features. Our findings show that the CONCH \cite{lu2024visual} and KEEP \cite{zhou2024knowledge} vision-language models can extract features with fewer center bias. Meanwhile, VIRCHOW-2 \cite{zimmermann2024virchow2} achieves the highest performance despite confounding features, suggesting that the scaling laws regarding dataset and model size promise foundation models obtaining a better understanding of the data. The comparison of the performance of both MIL settings highlights that attention-based MIL approaches are more robust against distribution shifts than prototype-based classifiers, which depend on the raw extracted features. 

Although our work covers an extensive range of histopathology FM, it is still limited to patch-level encoders. Future work could focus on benchmarking weakly supervised foundation models proposed to learn slide feature representations in an unsupervised fashion. Our benchmark relies on a multi-center dataset, which allows us to evaluate the presence of scanner-related information in the features extracted by a foundation model. Our proposed FM-SI score provides a quantitative and intuitive measurement of center-shift in the aggregated latent space of each encoder without requiring class labels. The insights about consistency to scanner shift presented in this work highlight the need for rigorous assessment of other distribution shifts, such as demographics-related ones, and their potential impact in real-world CPath applications. 

%
%

\section*{Acknowledgments}
We gratefully acknowledge the support from the Generalitat Valenciana (GVA) with the donation of the DGX A100 used for this work, an action co-financed by the European Union through the Operational Program of the European Regional Development Fund of the Comunitat Valenciana 2014-2020 (IDIFEDER/2020/030). This work has received funding from the Spanish Ministry of Economy and Competitiveness through the projects PID2022-140189OB-C21 (ASSIST) and CIPROM/2022/20 (COMTACTS2).

\bibliographystyle{splncs04}
\bibliography{references}

\newpage
\section*{Supplementary material}
\renewcommand{\thetable}{S\arabic{table}}
\setcounter{table}{0}
\renewcommand{\thefigure}{S\arabic{figure}}
\setcounter{figure}{0}

\subsection*{Robustness Index (RI)}
The medical center robustness index (RI) was proposed in \cite{de2025current} aiming to quantify how much biological and confounding features prevail in the outputs of histopathology foundation models. For a given sample ($i$), it measures the ratio between the number of samples that belong to the same biological class ($y$) and the number that represents the same medical center ($c$) in constrained neighborhood of $k$-samples. In our experiments we define, $k=25$. In summary, the RI for a given number of $k$-nearest samples ($RI_k$) is calculated as:

\begin{equation}
\label{eq_wt}
RI_k = \frac{\sum_{i=1}^{n} \sum_{j=1}^{k} \mathbf{1}(y_j = y_i)}{\sum_{i=1}^{n} \sum_{j=1}^{k} \mathbf{1}(c_j = c_i)}
\end{equation}

where \textbf{1}(·) is an indicator function: 1 if the condition is true, 0 otherwise.

\subsection*{Classification and bias-related metrics}
We provide in Table S1 a detailed summary of the performance metrics in the classification task for each foundation model (FM). We report the balanced accuracy for Attention-Based MIL (ABMIL) \cite{ilse2018attention} and Multiple Instance SimpleShot (MI-SimpleShot) \cite{chen2024towards}. For each FM, we provide the average BACC and standard deviation (in \%) through the 5 five folds in the designed cross-validation setting.

Regarding the center-shift performance measure, we include the score of the Robustness Index (RI) \cite{del2025fusocelular} and the proposed Foundation Model - Silhouette Index (FM-SI). The rows for each encoder in Table S1 are ordered in descending order of the FM-SI score. We highlight as the top-performer for FM-SI the model with the lowest score, thus suggesting poorer clustering of the data points regarding the center acquisition label. 

\begin{table}[]
    \label{tbl:sup}
    \caption{Classification performance and center-shift measurement metrics across the evaluated foundation models. Top-performing models are highlighted in \textbf{bold}. FM-SI: Foundation Model - Silhouette Index. RI: Robustness Index.}
    \centering
    \setlength{\tabcolsep}{4pt}
    \begin{tabular}{lcccc}
        \hline
        & \multicolumn{2}{c}{\textbf{Classification metrics}} & \multicolumn{2}{l}{\textbf{Center-shift metrics}} \\ \hline
        \textbf{Foundation model} & \textbf{ABMIL \cite{ilse2018attention}} & \textbf{MI-SimpleShot \cite{chen2024towards}} & \textbf{FM-SI (ours)} & \textbf{RI \cite{de2025current}} \\ 
        \hline
        PLIP \cite{huang2023visual}            & $67.53$\begin{tiny}$\pm5.14$\end{tiny} & $59.86$\begin{tiny}$\pm3.48$\end{tiny} & 0.6857 & 0.5790 \\
        GPFM \cite{ma2024towards}              & $80.88$\begin{tiny}$\pm4.16$\end{tiny} & $62.13$\begin{tiny}$\pm3.64$\end{tiny} & 0.6210 & 0.5763 \\
        UNI \cite{chen2024towards}             & $80.77$\begin{tiny}$\pm6.74$\end{tiny} & $68.66$\begin{tiny}$\pm3.01$\end{tiny} & 0.5973 & 0.6108 \\
        MUSK \cite{xiang2025vision}            & $81.72$\begin{tiny}$\pm5.37$\end{tiny} & $71.92$\begin{tiny}$\pm5.13$\end{tiny} & 0.3067 & 0.6945 \\
        VIRCHOW-2 \cite{zimmermann2024virchow2}& $\textbf{86.81}$\begin{tiny}$\pm\textbf{3.98}$\end{tiny} & \textbf{$\textbf{77.83}$\begin{tiny}$\pm\textbf{6.50}$\end{tiny}} & 0.3011 & 0.7848 \\
        CHIEF \cite{wang2024pathology}         & $80.77$\begin{tiny}$\pm6.76$\end{tiny} & $62.50$\begin{tiny}$\pm3.89$\end{tiny} & 0.2798 & 0.6200 \\
        CONCH \cite{lu2024visual}              & $83.00$\begin{tiny}$\pm3.82$\end{tiny} & $72.40$\begin{tiny}$\pm5.39$\end{tiny} & 0.0966 & 0.8040 \\
        KEEP \cite{zhou2024knowledge}          & $81.77$\begin{tiny}$\pm6.79$\end{tiny} & $72.91$\begin{tiny}$\pm5.91$\end{tiny} & \textbf{0.0283} & \textbf{0.8382} \\
        \hline
    \end{tabular}
\end{table}

\end{document}